\documentclass[letterpaper, 10 pt, conference]{ieeeconf}  

\IEEEoverridecommandlockouts                              

\title{\LARGE \bf
V2XP-ASG: Generating Adversarial Scenes for Vehicle-to-Everything Perception
}

\author{Hao Xiang$^{1*}$, Runsheng Xu$^{1*}$,Xin Xia$^{1}$, Zhaoliang Zheng$^{1}$, Bolei Zhou$^{1}$, Jiaqi Ma$^{1\dagger}$
\thanks{$^*$Equal contribution.}
\thanks{$^{1}$ University of California, Los Angeles, CA, USA.}
\thanks{$\dagger$ Corresponding author: \texttt{jiaqima@ucla.edu}.}
}

\usepackage[utf8]{inputenc} 
\usepackage[T1]{fontenc}    
\usepackage[colorlinks,linkcolor=black,
urlcolor=blue]{hyperref}
\usepackage{url}            
\usepackage{booktabs}       
\usepackage{amsfonts}       
\usepackage{nicefrac}       
\usepackage{microtype}      
\usepackage{xcolor}         
\usepackage{amsmath}

\DeclareMathOperator*{\argmin}{arg\,min}
\usepackage{bm}
\usepackage{algorithm}
\usepackage{algpseudocode}
\usepackage{multirow}
\usepackage{diagbox}
\usepackage{mathtools}
\usepackage{subcaption}
\usepackage{colortbl}
\usepackage{wrapfig}
\usepackage{makecell}
\usepackage{pifont}%
\usepackage{xcolor}

\definecolor{gray}{rgb}{0.5, 0.5, 0.5}
\colorlet{lightgray}{gray!20}

\begin{document}

\maketitle
\thispagestyle{empty}
\pagestyle{empty}

\begin{abstract} Recent advancements in Vehicle-to-Everything communication technology have enabled autonomous vehicles to share sensory information to obtain better perception performance. With the rapid growth of autonomous vehicles and intelligent infrastructure, the V2X perception systems will soon be deployed at scale, which raises a safety-critical question: \textit{how can we evaluate and improve its performance under  challenging traffic scenarios before the real-world deployment?} Collecting diverse large-scale real-world test scenes seems to be the most straightforward solution, but it is expensive and time-consuming, and the collections can only cover limited scenarios. To this end, we propose the first open adversarial scene generator V2XP-ASG that can produce realistic, challenging scenes for modern LiDAR-based multi-agent perception systems. V2XP-ASG  learns to construct an adversarial collaboration graph and simultaneously perturb multiple agents' poses in an adversarial and plausible manner. The experiments demonstrate that V2XP-ASG can effectively identify challenging scenes for a large range of V2X perception systems. Meanwhile, by training on the limited number of generated challenging scenes, the accuracy of V2X perception systems can be further improved by 12.3\% on challenging and 4\% on normal scenes. Our code will be released at \url{https://github.com/XHwind/V2XP-ASG}.
\end{abstract}

\section{Introduction}
Over the past decade, we have witnessed tremendous progress in Autonomous Vehicles~(AVs)~\cite{liu2020vision,liu2018intelligent,pal2021learning,khalil2022licanet,kitajima2022nationwide,papathanasopoulou2022data,betz2022autonomous,nayak2022uncertainty,masmoudi2021reinforcement,mukherjee2021predicting,christensen2021autonomous,de2021universally,liu2022yolov5,han2021auto} and intelligent transportation systems~\cite{xu2021opencda,chen2023dynamic,hua2019hierarchical}. Sooner, these autonomous systems will be deployed on roads at scale, opening up opportunities for cooperation between them.
 Previous works~\cite{yu2022dair, yuan2023generating,valiente2022robustness,ali2021optimizing,isprs-archives-XLIII-B2-2021-255-2021,xie2022safe} have demonstrated that by leveraging the Vehicle-to-Everything~(V2X) communication technology, AVs and infrastructure can perform cooperative perception by using the shared sensing information and thus significantly enhance the perception performance~\cite{xu2022v2x, hu2022where2comm,li2022v2x,yuan2022leveraging,lei2022latency}.
Despite the remarkable improvement, these works evaluate the proposed systems on the dataset with natural scenarios that do not contain sufficient safety-critical scenes. Under the challenging scenes, these systems may have inferior performance, and thus it is crucial to identify challenging scenes to fully understand the robustness of existing cooperative perception systems. The straightforward solution is to collect a wide range of testing scenes in the real world and identify critical ones. However, compared to single-agent systems, the cost and time consumption of gathering and labeling data for multi-agent systems can be much more demanding. A preferable cost-effective solution is to generate large-scale realistic scenes~\cite{tan2021scenegen,suo2021trafficsim,luo2020simulating} in high-fidelity simulators. Yet these approaches only consider the common scenes and lack the capability of performing stress tests on corner cases for the target systems. 



To address this problem, inspired by recent works~\cite{abeysirigoonawardena2019generating,rempe2022strive, wang2021advsim} that produce safety-critical scenarios for the single-agent planner by perturbing surrounding vehicles' trajectories, we aim to automatically generate diverse and challenging scenes for V2X perception systems. Nonetheless, it is demanding to directly transfer the single-agent scenario generation methods to V2X systems. Unlike the single-agent system with a single viewpoint, V2X system involves collaborators with multiple viewpoints. 
 The geometric relationships between these views from distinct collaborators can vastly influence the perception performance~\cite{liu2020who2com}, which is not considered in the previous works. 
Furthermore, the collaborator choices can influence the viewpoints and thus are coupled with the agent pose perturbations, leading to more complicated interactions. Handling these differences between single-agent and cooperative perception is the key to an effective V2X adversarial scene generator. However, no literature has been reported to investigate them. Moreover, most existing methods are tailored for planning, and there is no publicly available scene generation framework for perception, which hinders the progress of developing efficient scene generation methods for modern single-agent and V2X perception systems.



\begin{figure*}[!t]
\vspace{0.5mm}
\centering
\footnotesize
\includegraphics[width=\textwidth]{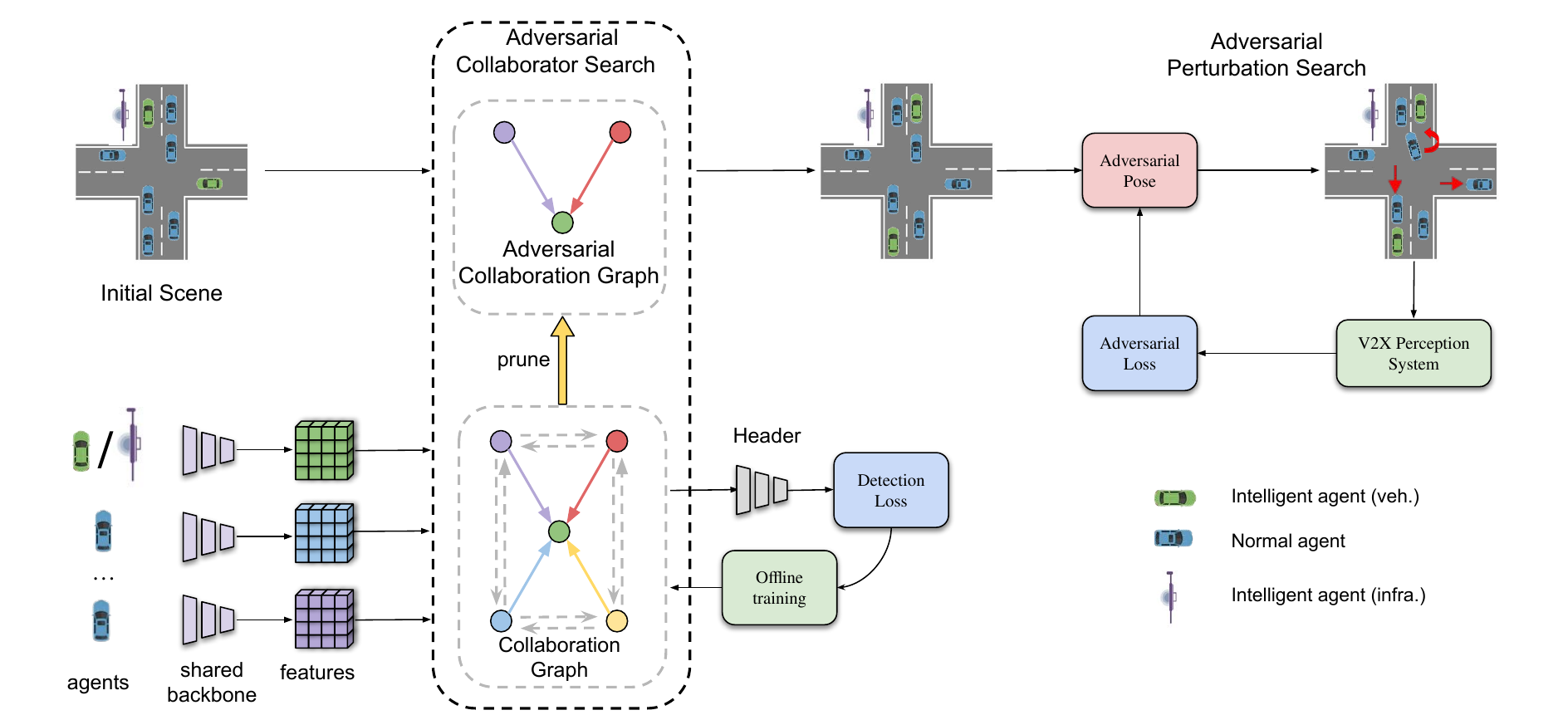}
\caption{\textbf{Overview of proposed V2XP-ASG.} It contains two steps: adversarial collaborator search and adversarial perturbation search. Details for each component can be found in Sec.~\ref{sec:method}}
\label{fig:framework}
\vspace{-4mm}
\end{figure*}

To this end, we introduce V2XP-ASG -- the first open \textbf{A}dversarial \textbf{S}cene \textbf{G}enerator for LiDAR-based \textbf{V2X} \textbf{P}erception systems. To the best of our knowledge, this is the first framework that aims to automatically generate challenging scenes for V2X perception. As shown in Figure~\ref{fig:framework}, given an initial scene from the dataset, V2XP-ASG first constructs an adversarial collaboration graph by searching for adversarial collaborators whose viewpoints combination will lead to inferior performance. Afterward, we perturb multiple agents' poses in an adversarial and plausible way. 
To reflect the updated LiDAR observation caused by  sensor viewpoint change and pose perturbation, we conduct the experiments based on the high-fidelity CARLA~\cite{dosovitskiy2017carla} simulator. Extensive experiments show that V2XP-ASG can create challenging scenes to severely deteriorate the performance of modern V2X perception systems. More importantly, training with these adversarial scenes can increase the system's accuracy.  Our main contributions can be summarized as follows
\begin{itemize}
    \item We present V2XP-ASG, the first open adversarial scene generation framework, to test modern V2X perception systems. Our V2XP-ASG can intelligently generate challenging scenarios for a wide range of V2X perception systems plausibly. We will release codes in the future.
    \item We formulate this scene generation task as a novel two-stage optimization problem, which makes the optimization easier and provides interpretability to the system's weakness. 
    \item We propose an efficient search strategy for the novel Adversarial Collaborator Search~(ACS) task in the first stage by leveraging learnable collaboration graph weights. Moreover, in the second stage, we employ the black-box optimization algorithms with customized adversarial objectives for perception tasks to perturb vehicle poses in an adversarial and feasible manner. These two stages are combined to efficiently raise the scene's challenging level.
    \item As demonstrated by our experiments, training on generated adversarial scenes can greatly improve the system's precision under both normal and challenging scenes.
\end{itemize}


\section{Related work}

\noindent\textbf{V2X Perception: }V2X perception investigates how to leverage the visual information from nearby AVs and intelligent infrastructure to enhance the perception capability. Based on the collaboration strategies, there are three major classes: early~\cite{chen2019cooper}, late~\cite{rawashdeh2018collaborative, rauch2012car2x, song2022vtc}, and intermediate fusion~\cite{wang2020v2vnet, xu2022opv2v, yuan2022keypoints, li2021learning, qiao2022adaptive, xu2022cobevt, cui2022coopernaut,Su2022uncertainty}. The early fusion method delivers the raw point clouds across agents, and each agent will feed the aggregated point clouds to the network for 3D detection. Despite preserving complete raw information, early fusion usually requires large bandwidth, making it unrealistic to deploy~\cite{song2022federated, song2022fedd3}. On the contrary, late fusion has a minimum data transmission size as it only circulates the metadata of prediction outputs. However, its accuracy is limited since it fails to provide valuable scenario context~\cite{wang2020v2vnet}. To achieve a good trade-off between bandwidth and accuracy, intermediate fusion, which broadcasts the compressed intermediate neural features, has been mostly studied recently. ~\cite{wang2020v2vnet} proposes a spatial-aware graph neural network for joint perception and prediction, and ~\cite{li2021learning} employs knowledge distillation to advance the learning with the supervision of early fusion. ~\cite{xu2022opv2v} proposes a location-wise self-attention mechanism to fuse the features from different AVs. This work evaluates all three fusion strategies and the single-agent perception method. 

\noindent\textbf{Adversarial Scenario Generation: }
As safety is critical to autonomous driving~\cite{9234108}, recently, several works~\cite{abeysirigoonawardena2019generating,rempe2022strive,wang2021advsim,klischat2019generating} have been proposed to generate safety-critical scenarios for identifying planning failures. ~\cite{abeysirigoonawardena2019generating} uses Bayes Optimization to search crash scenarios based on a hierarchical graph route in CARLA~\cite{dosovitskiy2017carla}. STRIVE~\cite{rempe2022strive} represents agents in latent space via a graph-based conditional VAE model and perturb latent variables via gradient-based optimization to produce trajectories that collide with a given planner. AdvSim~\cite{wang2021advsim} benchmarks several black-box optimization algorithms to search adversarial trajectories to obtain safety-critical scenarios for the full autonomy stack, but its adversarial objective is targeted for planning, and it is designed for the single vehicle system. Another stream of work~\cite{norden2019efficient,o2018scalable} formulates the scenario generation problem as the rare event simulation to sample failure scenarios for the single-agent autonomy system. In contrast, we focus on producing challenging scenarios for the LiDAR-based multi-agent V2X perception system where both the agents' poses and the selection of collaborators are searched to optimize an adversarial objective customized for perception.

\noindent\textbf{LiDAR-based adversarial attack: } LiDAR-based adversarial attacks~\cite{tu2020physically,li2021fooling,cao2019adversarial} in a physically realizable way have gained increasing attention.~\cite{cao2019adversarial} performs the first security study of LiDAR-based perception in AV settings by changing raw LiDAR points via a LiDAR spoofer.~\cite{li2021fooling} studies the backdoor of motion compensation and performs adversarial spoofing of the agent's trajectory to deteriorate the perception performance. In~\cite{tu2020physically}, an adversarial mesh is put on the rooftop of a vehicle, and mesh poses are optimized to make the vehicle invisible. Besides the above single-agent attack, ~\cite{tu2021adversarial} attacks the V2X perception system by adding adversarial noises to the intermediate features shared by intelligent agents. Different from previous work, we directly attack the agent collaboration choice and vehicle poses to deteriorate the performance of the modern V2X perception system, which ensures the feasibility and plausibility of the generated adversarial examples.

\noindent\textbf{Multi-agent collaboration graph: }
A classical multi-agent collaboration graph focuses on fusing information from selected agents or all the connected agents to increase the system's performance. Who2com \cite{liu2020who2com} proposes a handshake communication mechanism where the neural network can learn and determine which two agents should compress relevant information needed for each stage. When2com \cite{liu2020when2com} constructed learning-based communication groups to learn to communicate and used an asymmetric attention mechanism to decide when to communicate across a fully-connected graph. DiscoGraph \cite{li2021learning} exploited a method that incorporates both early and intermediate collaboration into a knowledge distillation framework, which enables the knowledge of early collaboration to guide the training of an intermediate fusion model. However, rather than trying to improve the perception performance by finding robust collaborators, in this work, we are interested in searching for adversarial collaborators to deteriorate the task performance, which can help expose the potential vulnerability of the multi-agent system. 
\section{Adversarial Scene Generation}
\label{sec:method}
V2XP-ASG aims to generate realistic, challenging scenes for a given V2X perception model.
The overall architecture of our framework is shown in Figure. \ref{fig:framework}, which consists of Adversarial Collaborator Search (ACS) and Adversarial Perturbation Search (APS). Given an existing scene with a fixed ego agent, we first search for adversarial collaborator combinations by leveraging an attention-based sampling approach and then build the adversarial collaboration graph with selected agents. Afterward, we perturb multiple agents' poses and feed updated LiDAR observation into the target perception system to generate 3D bounding boxes. Eventually, we evaluate the predictions with adversarial objectives and perform black-box optimization to update poses. 
\subsection{Problem formulation}
Given an initial scene with a set of agents $\mathcal{A}=\{a_1,\dots,a_N\}$ where agents $a_i$ can be either vehicle or infrastructure, we assume $k$ of them are equipped with sensing and communication devices and are denoted as intelligent agents $\mathcal{I}\subset\mathcal{A}$. Within the communication range, all the agents belonging to $\mathcal{I}$ can share information with each other and one of them is selected as the ego agent to aggregate the sensing information to form a unified detection. More formally, the set of observations from these agents are
\begin{equation}
    \mathcal{X}\left(\mathcal{S}, \mathcal{I}\right) = \{\mathbf{X}_i\left(\mathcal{S}\right) | \forall i\in \mathcal{I}\}\vspace{-1mm}
\end{equation}
where $\mathcal{S}=\{\mathbf{s}_1,\dots,\mathbf{s}_N\}$ is the set of poses of all agents in the scene. $\mathbf{X}_i\left(\mathcal{S}\right)$ is the sensing observation from intelligent agent $i$ and it reflects the change of agents' poses $\mathcal{S}$. The ego agent with neighboring collaborators will collaboratively predict the bounding boxes $\hat{Y}=f\left(\mathcal{X}\left(\mathcal{S}, \mathcal{I}\right);\tau\right)$ based on the aggregated sensing observations $\mathcal{X}$, where $f$ is the V2X 3D detection model, and $\tau$ is the fusion strategy including early , late and intermediate fusion.

Our objective is to attack the V2X perception model by adjusting initial normal scenes to be more challenging. There are two types of adversaries in this work, namely agent collaboration combination $\mathcal{I}$ and pose perturbation $\Delta$. These two adversaries will modify the sensor viewpoint and vehicle poses in a physically plausible way and CARLA~\cite{dosovitskiy2017carla} is leveraged to simulate the updated observations $\mathcal{X}\left(\mathcal{S}+\Delta, \mathcal{I}\right)$.

We then define the adversarial objective $\mathcal{L}_{\text{adv}}$ (Section \ref{sec:obj}) which is minimized to generate challenging scenes: 
\begin{equation}
\begin{aligned}
\Delta^*,\mathcal{I}^*=\argmin_{\Delta,\mathcal{I}} \quad & \mathcal{L}_{\text{adv}}\left(f\left(\mathcal{X}\left(\mathcal{S}+\Delta, \mathcal{I}\right);\tau\right), Y\right)\\
\textrm{s.t.} \quad&\|\Delta\|\le \delta, \mathcal{I}\subset \mathcal{A}, |\mathcal{I}|=k\\
\end{aligned}
\end{equation}
where $|\cdot|$ is the set cardinality and $Y$ is the ground truth bounding boxes. We decouple the above formula and optimize this factorized approximation as follows: 
\begin{align}
\mathcal{I}^*=\argmin_{\mathcal{I}} \quad & \mathcal{L}_{\text{adv}}\left(f\left(\mathcal{X}\left(\mathcal{S}, \mathcal{I}\right);\tau\right), Y\right)\nonumber\\
\textrm{s.t.} \quad&\mathcal{I}\subset \mathcal{A}, |\mathcal{I}|=k \label{eq:optim1}\\
\Delta^*=\argmin_{\Delta} \quad & \mathcal{L}_{\text{adv}}\left(f\left(\mathcal{X}\left(\mathcal{S}+\Delta, \mathcal{I}^*\right);\tau\right), Y\right)\nonumber\\
\textrm{s.t.} \quad&\|\Delta\|\le \delta \label{eq:optim2}
\end{align}
Such factorization can separate optimization into two subsequent tasks: adversarial collaborator search (Eq.~\ref{eq:optim1}) and adversarial pose perturbation (Eq.~\ref{eq:optim2}), which can not only simplify the optimization but also provides the possibility of analyzing two stages individually to better understand the performance of the target system. As the performance drop brought by pose perturbation can be easily diminished by changing sensor viewpoints, we choose to search collaboration first instead of the reverse order. For multi-agent systems, $k=1$ will clearly lead to inferior performance. However, we are interested in studying the robustness of V2X system even when it is deployed at scale. Thus we fix $k=3$ in this work. 



\subsection{Adversarial Collaborator Search}
\label{sec:acs}

 
 
Although examining all the agent combinations may seem appealing at first but it is prohibitively expensive due to the enormous combinations and the high computation cost. Thus instead, we design an efficient search algorithm. We adopt the intermediate fusion model AttFuse~\cite{xu2022opv2v} to construct the collaboration graph where the learnable edge weights are considered to represent the importance of that agent's feature contribution to the ego agent. We leverage these learnable edge weights to define each agent's weakness level so that lower values represent smaller contributions to the overall perception system and thus, assigning these weak agents with higher sampling probability can increase the chance of finding adversarial collaborators with inferior performance. Although other fusion methods such as Early Fusion and Late Fusion can't produce collaboration graphs with important weights, we experimentally demonstrate that the proposed attention-based probabilistic sampling method can be transferred to other models with great search efficiency (~\ref{tab:acs}). In addition, our framework is general and other adversarial collaborator search methods can be integrated into the framework and we encourage more researchers to investigate this new direction.


 To obtain the edge weights for all agents, we first equip all agents with LiDAR sensor in the simulator. Each agent will reason intermediate features based on its raw sensing observation and then transmit their intermediate features $\mathbf{H}_{i}\in\mathbb{R}^{H \times W \times C}$ to the ego agent. Then the self-attention model is employed to capture the feature importance from different agents in the same spatial locations under ego coordinate. Formally, let $\mathbf{h}_{mn}^i=[\mathbf{H}_{i}]_{mn}\in\mathbb{R}^{C}$ be the feature from agent $i$ in location $(m,n)$ and $\mathbf{h}_{mn}=\{\mathbf{h}_{mn}^1,\dots, \mathbf{h}_{mn}^N\}\in\mathbb{R}^{N\times C}$ is the aggregated features for location $(m,n)$ from all $N$ agents. Then the attention is operated as follows:
\begin{align}
    \mathbf{a}_{mn} &= \text{softmax}\left(\frac{\mathbf{q}_{mn} \mathbf{k}_{mn}^T}{\sqrt{C}}\right)\vspace{-1mm}
\end{align}
where $\mathbf{q}_{mn}, \mathbf{k}_{mn}$ and $\mathbf{v}_{mn}$ are linear projections of $\mathbf{h}_{mn}$ along channel dimension. The fused feature $\mathbf{h'}_{mn}=\mathbf{a}_{mn}\mathbf{v}_{mn}$. It can be sliced and rearranged to get the updated feature $\mathbf{H'}_{i}$. 

Before applying this self-attention model in our scene generator, we will train it on the augmented OPV2V dataset~\cite{xu2022opv2v} to learn the collaboration graph construction by sending  $\mathbf{H'}_{i}$ to a detection header to produce bounding box predictions.  During the adversarial procedure, we fix the network's weights and directly exploit  calculated $ \mathbf{a}_{mn}$ to detect the adversarial collaborators. As the attention weights are position-wise, average pooling is adopted to aggregate the weights of all spatial locations. 
\begin{equation}
s_{j} = \text{Avg}\left([\mathbf{a}_{11}]_{i_*j},\dots,[\mathbf{a}_{HW}]_{i_*j}\right)\vspace{-1mm}
\end{equation} 
Here $s_{j}$ represents the importance of agent $j$ and thus $1/s_{j}$ is the associated weakness level for this agent. For each collaborator combination $\mathcal{I}$, we can model its weakness level as the sum of the individual agent's weakness level and apply softmax to form a probability distribution: 
\begin{align}
\mathbf{w} &= \{w_{\mathcal{I}}=\sum_{i\in\mathcal{I}}\frac{1}{s_{i}} | \mathcal{I}\subset\mathcal{A}, |\mathcal{I}|=k\}\\
    \mathbf{p}&=\text{Softmax}\left(\frac{\mathbf{w}}{\tau}\right)\label{eq:prob}
\end{align}
where $\tau$ is the temperature parameter to control the shape of the distribution. We sample $k_0$ combinations without replacement according to the probability $\mathbf{p}$, examine all the adversarial objective values, and keep the best one with the lowest adversarial value as the adversarial collaboration combination $\mathcal{I}^*$. This combination is eventually used to build the adversarial collaboration graph where only the selected agents can share the information with each other. 
\begin{figure}
\vspace{-2mm}
\begin{algorithm}[H]
    \caption{\textbf{Adversarial perturbation search}}\label{alg:aps}
    \begin{algorithmic}[1]
    \State Sample $m$ agents to perturb according to heuristics
    \State Generate context-aware feasible set $\mathcal{Q}$ for $m$ agents
    \State Initialize historical observation $\mathcal{D}=\{\}$
        \For{$i=1\dots M$}
            \State Generate $\Delta^{(i)}$ based on historical observation $\mathcal{D}$ and black-box search algorithm
            \State Project $\Delta^{(i)}$ onto feasible set $\mathcal{Q}$ 
            \State Get updated sensing observation from LiDAR simulator $\mathcal{X}^{(i)}=\mathcal{X}\left(\mathcal{S}+\Delta^{(i)}, \mathcal{I}^*\right)$
            \State $\mathcal{L}_{\text{adv}}^{(i)}=\mathcal{L}_{\text{adv}}\left(f\left(\mathcal{X}^{(i)}\right), Y\right)$
            \State $\mathcal{D}=\mathcal{D}\cup\{(\Delta^{(i)}, \mathcal{L}_{\text{adv}}^{(i)})\}$
        \EndFor
        \State $\Delta^{(*)}=\argmin_{\Delta^{(i)}, i\in\{1,\dots,M\}} \mathcal{L}_{\text{adv}}^{(i)}$
    \end{algorithmic} 
\end{algorithm}
\vspace{-8mm}
\end{figure}
\subsection{Adversarial Perturbation Search}
\label{sec:aps}
\noindent\textbf{Search space: }We perturb multiple agents' poses within physical plausible bounds. Each agent's perturbation is parameterized as $\bm{\delta_i}=\left(\delta x_i, \delta y_i, \delta \theta_i\right)$,  where $\left(\delta x_i, \delta y_i\right)$ is the perturbation of the location and $\delta \theta_i$ is the change of yaw angle. The multi-agent perturbation is then 
\begin{align}
    \Delta = \left\{\bm{\delta_1}, \bm{\delta_2}, \dots, \bm{\delta_m}\right\}
\end{align}
where $m$ is the number of perturbed agents and these $m$ agents are sampled according to an occlusion-inspired heuristic from the set $\mathcal{A}$. We rank $N$ agents by occlusion levels in decreasing order and then select the top $m$ agents to perturb. Specifically, each agent's occlusion level is the sum of its intrinsic-occlusion score and extrinsic-occlusion score. The intrinsic-occlusion score is 1 if the agent is partially overlapped by any other agent in the scene, indicating the occlusion level of the perturbed agent (occludee). The extrinsic-occlusion score measures how many other agents in the scene can be partially occluded by the examined agent (occluder). For the multi-agent V2X system, we sum the occlusion score from different viewpoints as the agent's overall score. In this way, the sampled agents are more likely to observe partially occluded objects.  We set $\|\Delta\|\le\delta$ to constrain the perturbation within a limited range. In this work, $m = 3$ is adopted for increased search space and scenario configurations. \\
\textbf{Black-box search algorithm: }Our framework is amenable to any black-box search algorithm. The search algorithm seeks to find challenging scenes for the V2X perception model by minimizing the adversarial objective $\mathcal{L}_{\text{adv}}$. In this paper, we benchmark 3 black-box search algorithms: Random Search (RS), Genetic Algorithm (GA)~\cite{alzantot2019genattack} and Bayesian Optimization (BO)~\cite{ru2019bayesopt}. The Random Search will randomly sample perturbation from the feasible set. The Genetic Algorithm will maintain a population of candidate perturbations, evolve the population over time by selecting parents with high fitness scores, and the best candidate is always kept for enhanced performance~\cite{alzantot2019genattack}. The Bayesian Optimization will build a surrogate model and use the acquisition function to balance the exploration and exploitation for generating candidates.

\textbf{Search pipeline: }The overall method is summarized in Algorithm \ref{alg:aps}. To increase the plausibility of the perturbation, we build a context-aware feasible perturbation set $\mathcal{Q}$ for the sampled $m$ agents. To generate such a feasible set, we first uniformly sample $N_{\mathcal{Q}}=1000$ potential perturbations within the bounds ($\delta x, \delta y\in[-2.5,2.5]$~m and $\delta \theta\in[-45^\circ, 45^\circ]$) and remove the ones that can cause potential collisions. As the location and angle have different magnitudes, to diminish this bias, we also normalize the perturbation $\Delta$ by dividing it by the perturbation range s.t. $\|\Delta\|_{\infty}\le1$. A historical observation $\mathcal{D}$ of $\left(\Delta, \mathcal{L}_{\text{adv}}\right)$ pairs is maintained through the optimization process. During each iteration, the search algorithm generates the potential perturbation $\Delta$ and projects it onto the feasible set $\mathcal{Q}$ by finding the closest element in the set measured in $\ell_2$ distance. Then it will query the target V2X perception model to obtain adversarial loss $\mathcal{L}_{\text{adv}}$. The search algorithm will update the historical observation $\mathcal{D}$, which is used to generate the perturbation for the next iteration. The best-observed perturbation with the lowest adversarial loss is selected for the final perturbation.  
\subsection{Adversarial Objective}
\label{sec:obj}
We adopt weighted average precision as the adversarial loss. 
\begin{equation}
    \mathcal{L}_{\text{adv}} = \sum_{t\in\mathcal{T}}w_t\text{AP@}t
\end{equation}
where  $\mathcal{T}$ is a set of Intersection of Union (IoU) thresholds and in our experiment, $\mathcal{T}=\{0.3,0.5,0.7\}$. $\text{AP@}t$ is the average precision (AP) at IoU of threshold $t$ and $w_t$ is the associated weight. Using this weighted sum of average precision can lead to smoother score change, which eases the difficulty of optimization. In this work, we set $w_{0.3}=1$, $w_{0.5}=0.8$ and $w_{0.7}=0.5$ to give higher weights for lower IoU thresholds as their APs are generally harder to decrease. 
\section{Experiments}

\subsection{Experiment Setup}
\noindent\textbf{Dataset: }Experiments are conducted in CARLA~\cite{dosovitskiy2017carla} simulator.
Two datasets are used in the experiment: 1) we augment the existing large-scale Vehicle-to-Vehicle (V2V) perception dataset, namely OPV2V, with infrastructure sensor data. The OPV2V's training split is used to train modern V2X perception models until convergence. From the augmented OPV2V's test split, we have 94 driving logs and each has 70 frames, covering 8 road areas. Due to the high computational cost of running the CARLA simulator and deep networks, We sample 331 \textit{Normal} scenes from these collected scenarios, and the train/validation splits for evaluating V2XP-ASG are 219/112. The major experiments are conducted in train split and the validation split is only used for reporting the performance of the fine-tuned model. 2) Besides the above data, we build another heuristic-generated challenging dataset called \textit{Heuristic} that has high traffic density with severe occlusions and sparse observations. This hold-out dataset includes 321 frames sampled from 14 driving logs.
We first collect 14 driving logs for a total of 980 frames with dense traffic (20$\sim$30 vehicles per frame) in 4 CARLA towns. Then for each frame, we calculate the average number of LiDAR points within the bounding boxes, and we curate the raw data by only keeping the scene with an average number of points less than the threshold $N_{thred}=25$. As this dataset is only visible during inference, it provides a fairer evaluation of whether the fine-tuned model can achieve better results in unseen challenging scenes (Table~\ref{tab:robust}).

\begin{table}[]
\vspace{1.6mm}
\footnotesize
\setlength{\tabcolsep}{2pt}
    \centering
    \caption{Evaluation of V2X perception models on \textit{Normal} and V2XP-ASG generated \textit{Challenging} scenes. The number within parentheses is the AP drop compared with \textit{Normal}. }
    \label{tab:main_table}
    \begin{tabular}{c|c|l|l|l}
         \cellcolor{lightgray} {Methods}&\cellcolor{lightgray} {Scenes} \cellcolor{lightgray} {Type}&\cellcolor{lightgray} {AP@0.3}&\cellcolor{lightgray} {AP@0.5}& \cellcolor{lightgray} {AP@0.7}\\
         \toprule
         \multirow{2}{*}{No Fusion}&Normal&55.4 & 54.8 & 46.3\\
         &Challenging&31.1(\textbf{-24.3}) & 30.3(\textbf{-24.5}) & 25.1(\textbf{-21.2})\\
         \midrule
         \multirow{2}{*}{Late Fusion}&Normal&73.4 & 72.6 & 62.6\\
         &Challenging& 42.0(\textbf{-31.4}) & 40.0(\textbf{-32.6}) & 32.8(\textbf{-29.8})\\
         \midrule
         \multirow{2}{*}{Early Fusion}&Normal&80.3 & 79.8 & 73.5\\
         &Challenging&49.8(\textbf{-30.5}) & 48.3(\textbf{-31.5}) & 43.4(\textbf{-30.1})\\
         \midrule
         \multirow{2}{*}{AttFuse}&Normal&82.4&81.5& 74.6\\
         &Challenging & 46.6(\textbf{-35.8}) & 44.9(\textbf{-36.6}) & 40.5(\textbf{-34.1})\\
         \bottomrule
    \end{tabular}
    \vspace{-6mm}
\end{table}

\begin{figure*}[!t]
\vspace{2mm}
\centering
    \begin{subfigure}[c]{0.21\linewidth}
        \centering{\includegraphics[width=1\linewidth]{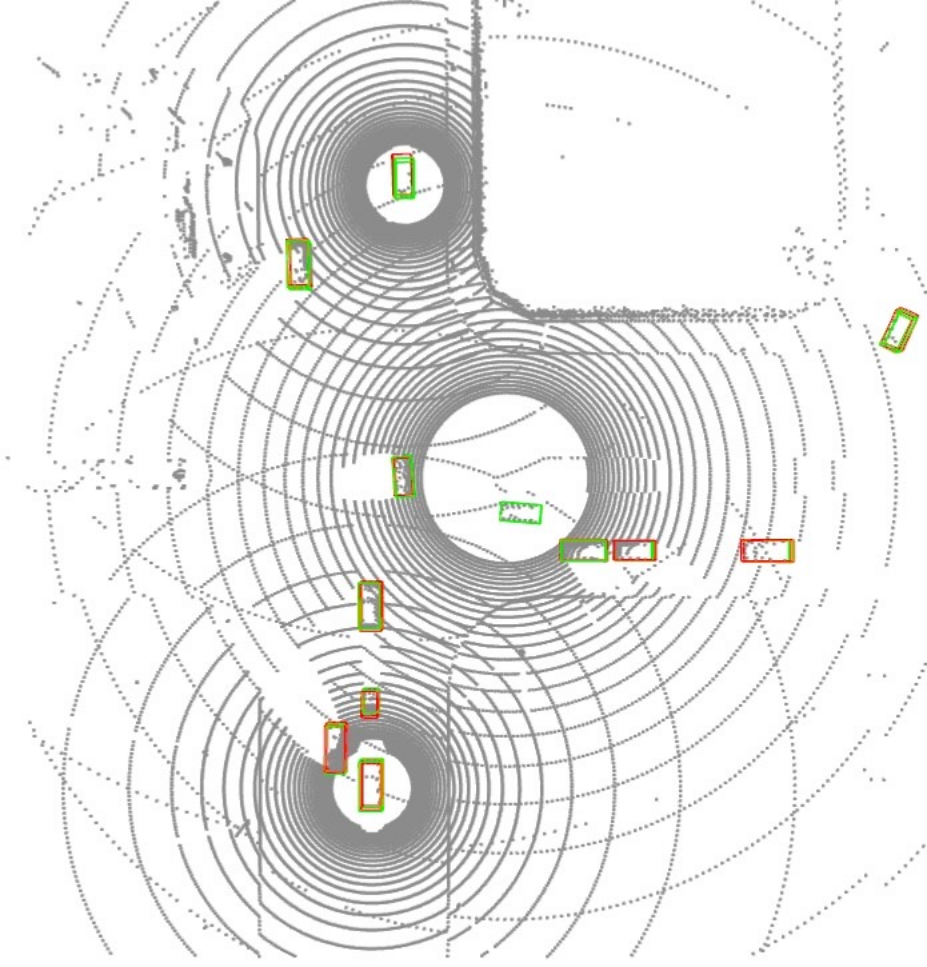}}
        \caption{Initial scene}
        \label{fig:qualitive-a}
    \end{subfigure}
    \begin{subfigure}[c]{0.21\linewidth}
        \centering{\includegraphics[width=1\linewidth]{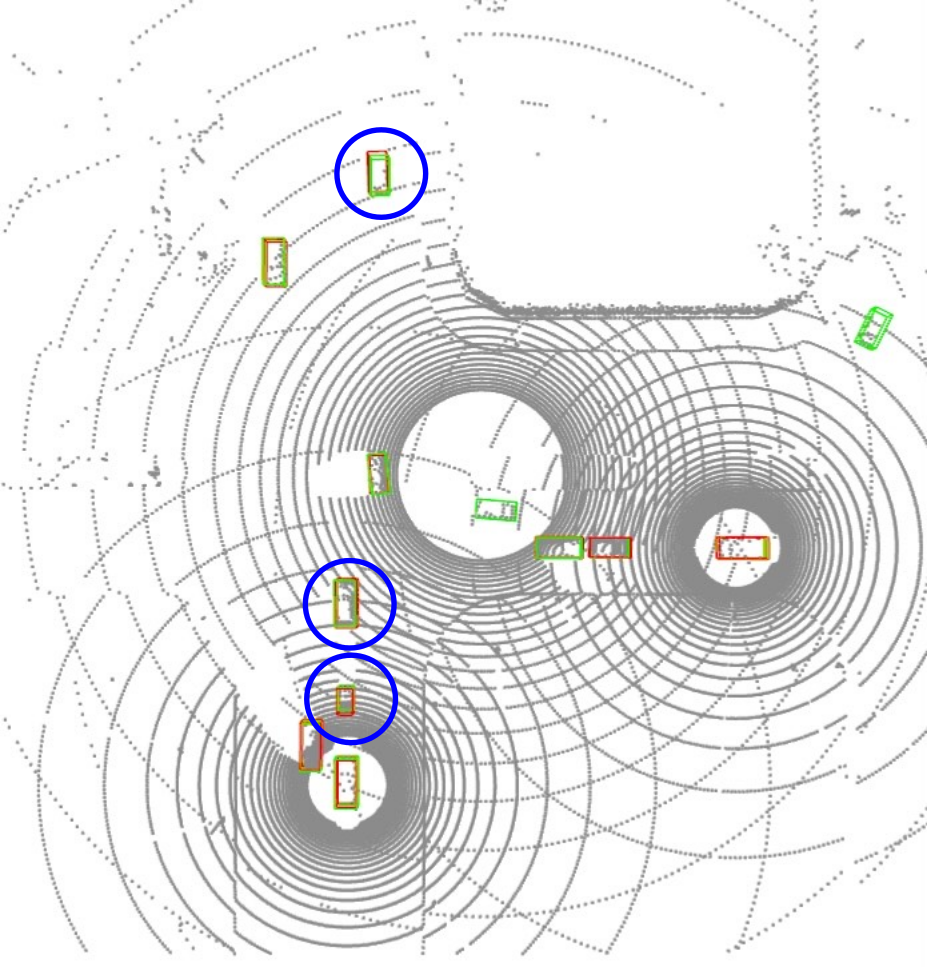}}
        \caption{After ACS}
        \label{fig:qualitive-b}
    \end{subfigure}
    \begin{subfigure}[c]{0.21\linewidth}
        \centering{\includegraphics[width=1\linewidth]{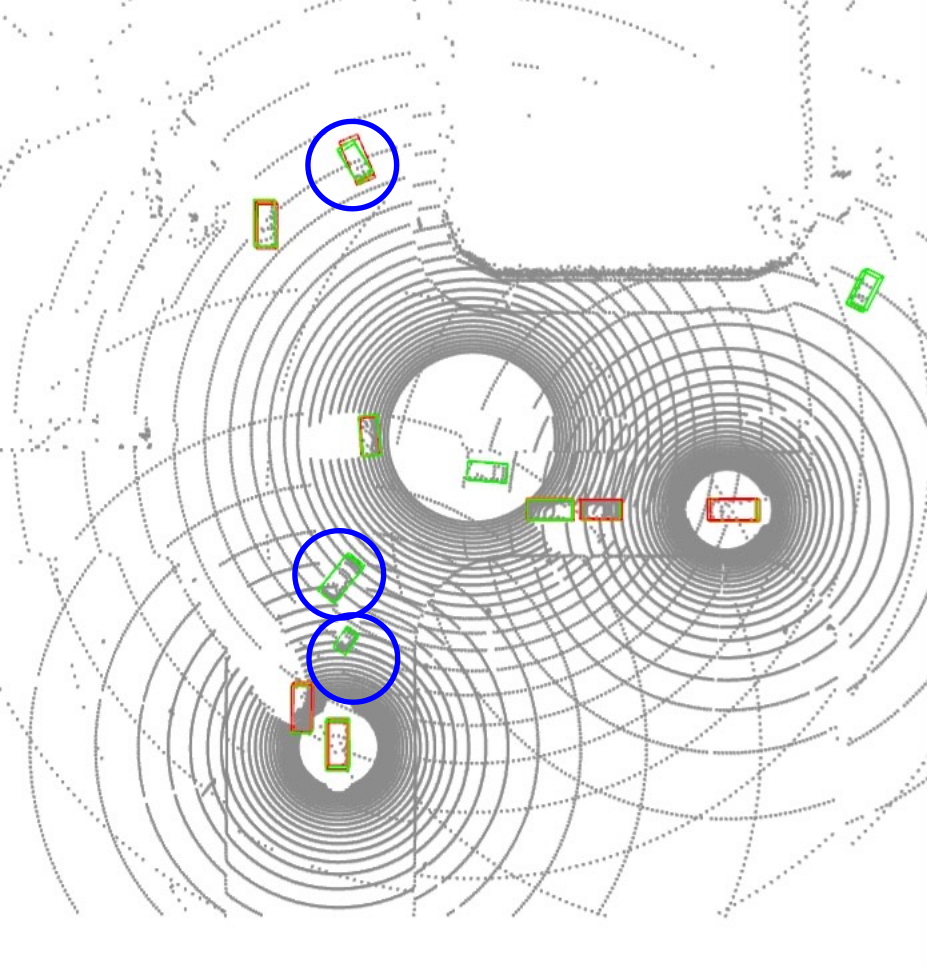}}
        \caption{Final scene}
        \label{fig:qualitive-c}
    \end{subfigure}
    \caption{Qualitative results of  generated challenging scenes. (a) initial scene. (b) scene after selecting adversarial collaborators. (c) scene after vehicle pose perturbation.  Blue circles represent the perturbed vehicles.}
    \label{fig:qualitive}
    \vspace{-3mm}
\end{figure*}

\begin{table*}[]
\centering
\footnotesize
\setlength{\tabcolsep}{2pt}
\begin{tabular}{@{}cccc@{}}
\begin{minipage}[t]{0.246\textwidth}
\centering
\setlength{\tabcolsep}{2pt}
\renewcommand{\arraystretch}{1.}
\caption{{Improving AttFuse study. FT means fine-tuning on train split of generated scenes}}
\label{tab:robust}
\begin{tabular}{c|c|c}
\cellcolor{lightgray}  {Scene Types} &
 \cellcolor{lightgray} {Models} &
 \cellcolor{lightgray} {AP@0.7}\\ \toprule
\multirow{2}{*}{Normal}&AttFuse& 70.6\\
         &AttFuse+FT& \textbf{74.6}\\
         \midrule
         \multirow{2}{*}{Heuristic}&AttFuse& 64.9\\
         &AttFuse+FT& \textbf{73.1}\\
         \midrule
         \multirow{2}{*}{Challenging}&AttFuse & 43.9\\
         &AttFuse+FT&  \textbf{56.3}\\
         \bottomrule
\end{tabular}
\end{minipage}
&
\begin{minipage}[t]{0.346\textwidth}
\centering
\setlength{\tabcolsep}{2pt}
\renewcommand{\arraystretch}{1.}
\caption{{Transferability experiment in AP@0.7.} The number in parentheses indicates the AP drop compared with \textit{Normal}.}
\label{tab:transferability}
\begin{tabular}{c|c|c|c}
         \cellcolor{lightgray}{}&\cellcolor{lightgray}{}&\cellcolor{lightgray}{}&\cellcolor{lightgray}{}\\
         \multirow{-2}{*}{\cellcolor{lightgray} {\diagbox{Target}{Source}}}&\multirow{-2}{*}{\cellcolor{lightgray} {Late}}&\multirow{-2}{*}{\cellcolor{lightgray} {Early}}&\multirow{-2}{*}{\cellcolor{lightgray} {AttFuse}}\\
         \toprule
         Late&\textbf{32.8(-29.8)} & 41.3(-21.3) & 39.8(-22.8)\\
         Early& 51.8(-21.7) & \textbf{43.4(-30.1)} & 49.4(-24.1)\\
         AttFuse& 48.1(-26.5) &  48.0(-26.6) &  \textbf{40.5(-34.1)}\\
         \bottomrule
    \end{tabular}
\end{minipage}
&
\begin{minipage}[t]{0.225\textwidth}
\centering
\setlength{\tabcolsep}{3pt}
\renewcommand{\arraystretch}{1.}
\caption{{Collaboration selection method comparison in AP@0.7.}}
\label{tab:acs}
\begin{tabular}{c|c|c|c}
         \cellcolor{lightgray} {}&\cellcolor{lightgray} {Late}&\cellcolor{lightgray} {Early}&\cellcolor{lightgray} {AttFuse} \\
         \toprule
         Normal&62.6&73.5&74.2  \\
         ACS-R&56.1&66.7&70.1 \\
         ACS-A&\textbf{53.1}&\textbf{64.6}&\textbf{66.5} \\
         \bottomrule
    \end{tabular}
\end{minipage}
&
\begin{minipage}[t]{0.15\textwidth}
\caption{{Component ablation study of ACS and APS.}}
\label{tab:component}
    \begin{tabular}{cc|c}
         \cellcolor{lightgray} {ACS}&\cellcolor{lightgray} {APS}&\cellcolor{lightgray} {AP@0.7}   \\
         \toprule
         & & 74.6\\
         \checkmark& &  66.5\\
         &\checkmark&44.4\\
         \checkmark&\checkmark&\textbf{40.5}\\
         \bottomrule
    \end{tabular}
\end{minipage}
\vspace{-3mm}
\end{tabular}

\end{table*}
\begin{table}[t]
    \centering
    \caption{Benchmark results of the black-box algorithms. }
    \label{tab:black_box}
    \begin{tabular}{c|c|c|c}
        \cellcolor{lightgray}{Methods}&\cellcolor{lightgray}{AP@0.3}&\cellcolor{lightgray}{AP@0.5}&\cellcolor{lightgray}{AP@0.7} \\
        \toprule
        Normal&82.4&81.5&74.6\\
         Ransom Search (RS)&68.3&67.4 & 60.4  \\
         Genetic Algorithm (GA)&54.8 & 52.6&46.8 \\
         Bayesian Optimization (BO)&50.9&49.4& 44.4 \\
         \bottomrule
    \end{tabular}
    \vspace{-4mm}
\end{table}

\noindent\textbf{V2X perception model: }V2XP-ASG is evaluated on 4 models: 1) {No Fusion}, which predicts bounding boxes only based on ego-agent LiDAR point clouds. 2) {Late Fusion}, where detection proposals of each agent are transmitted and the non-maximum suppression is applied to generate the final predictions. 3) {Early Fusion}, which aggregates raw LiDAR points from different agents to a holistic view and feeds them to the detector. 4) {AttFuse}~\cite{xu2022opv2v}, an intermediate fusion model which broadcasts intermediate features with each other and uses location-wise self-attention to fuse received features. All the models are implemented with PointPillar~\cite{lang2019pointpillars} backbone. 

\noindent\textbf{V2XP-ASG: }We adopt~{AttFuse}'s attention weights for constructing the adversarial collaboration graph for all 3 V2X perception models. The temperature parameter $\tau=0.03$ and we examine $k_0=3$ combinations sampled according to  Eq.~\ref{eq:prob}. For the main experiments, we adopt the Bayesian Optimization~\cite{ru2019bayesopt} as the black-box search algorithm. For LiDAR simulations, to diminish the influence of noise, we set the noise to be zero i.e., zero drop-off rate and zero noise level so that the perception performance is solely affected by the intrinsic scene configurations. The perception performance is evaluated in the range of $x,y\in[-48,48]~$m. For consistent ground truth, when perturbing agents, we ensure that the updated poses still stay within the evaluation range. Parameters of black-box optimization can be found in the supplementary videos.

\subsection{Experimental results}
\textbf{V2X perception model evaluations: }The evaluation results for different V2X perception models are shown in Table~\ref{tab:main_table}. The results demonstrate that our V2XP-ASG can generate challenging scenes for both single-agent and V2X perception systems with an average of 28.7\% drop of AP@0.7, showing that V2XP-ASG can efficiently identify challenging scenes for a wide range of models with different fusion strategies. 


\textbf{Improvement with challenging scenes: }We now investigate if our generated challenging scenes can help improve the performance of the V2X perception system. We first fine-tune the original AttFuse model on train split of its generated challenging scenes and then test the model on the val split of \textit{Normal} and \textit{Challenging} scenes. For a fairer comparison, we also evaluate its performance on a separate challenging hold-out set (\textit{Heuristic}). As shown in Table~\ref{tab:robust}, the fine-tuned model show improvement for all 3 datasets, illustrating the great benefit of V2XP-ASG for improving perception systems.

\textbf{Transferability of challenging scenes: }Table~\ref{tab:transferability} shows the transferability of V2XP-ASG generated scenes. The source model is used for generating challenging scenes and the target model is tested on these generated scenes. The performance is the best when testing the target model on the scenes generated by the same model. Moreover, three models all show great performance drop on each other's generated challenging scenes than their performance on normal scenes, showing that the models have a certain degree of agreement on whether a scene is challenging and thus demonstrating the favored transferability of V2XP-ASG generated challenging scenes.

\textbf{Baseline comparison: }We compare the proposed AttFuse-based collaborator selection strategy (ACS-A) with random selection (ACS-R) that randomly sample $k_0=3$ agent combinations. The experiment results are shown in Table~\ref{tab:acs}. The proposed method can decrease the detection performance by 9.5\%, 8.9\%, 7.7\% while random selection can only decrease the score by 6.5\%, 6.8\%, 4.1\%, showing great efficiency of the proposed search strategy. 

\textbf{Component analysis: } As shown in Table~\ref{tab:component}, the proposed two components ACS and APS are both instrumental for finding challenging scenes, demonstrating that both viewpoints and pose perturbation are critical for identifying challenging scenes for the multi-agent system.  

\textbf{Black-box search benchmark results: }The benchmark results for different black-box search strategies are shown in Table~\ref{tab:black_box}. Both GA and BO outperform RS and BO reaches the best performance. We argue this is due to the good balance between exploration and exploitation of BO.

\textbf{Visualization results: }
Figure~\ref{fig:qualitive} shows the visualization of V2XP-ASG generated \textit{Challenging} scenes and the associated \textit{Normal} scenes. As shown in Figure~\ref{fig:qualitive-b}, adversarial collaborator selection can change sensor viewpoint  to make objects hard to detect. In Figure~\ref{fig:qualitive-c}, three agents are perturbed and two of the perturbed agents simultaneously take a right turn, which is realistic and rare in real world. It demonstrates that the V2XP-ASG generated scenes are both challenging and meaningful for testing the perception module.

\section{Conclusions}
In this work, we present the first open adversarial scene generation framework dubbed V2XP-ASG for LiDAR-based V2X perception systems. The experiments demonstrate that the V2XP-ASG can create challenging scenes for a wide range of perception models, including single-agent and V2X perception systems. In particular, by training on generated scenes, the performance of the perception system can be further improved for both common and challenging scenes. Moreover, the ablation study verifies that the proposed attention-based probabilistic sampling approach can effectively find adversarial collaborators and the two novel components are both advantageous for generating challenging scenes. 

\section{Acknowledgement}
This work is part of the OpenCDA Ecosystem~\cite{10045043} and is supported by the  Federal Highway Administration with Grant number 693JJ321C000016.







\bibliographystyle{IEEEtran}
\bibliography{IEEEfull}

\begin{thebibliography}{10}
\providecommand{\url}[1]{#1}
\csname url@samestyle\endcsname
\providecommand{\newblock}{\relax}
\providecommand{\bibinfo}[2]{#2}
\providecommand{\BIBentrySTDinterwordspacing}{\spaceskip=0pt\relax}
\providecommand{\BIBentryALTinterwordstretchfactor}{4}
\providecommand{\BIBentryALTinterwordspacing}{\spaceskip=\fontdimen2\font plus
\BIBentryALTinterwordstretchfactor\fontdimen3\font minus
  \fontdimen4\font\relax}
\providecommand{\BIBforeignlanguage}[2]{{%
\expandafter\ifx\csname l@#1\endcsname\relax
\typeout{** WARNING: IEEEtran.bst: No hyphenation pattern has been}%
\typeout{** loaded for the language `#1'. Using the pattern for}%
\typeout{** the default language instead.}%
\else
\language=\csname l@#1\endcsname
\fi
#2}}
\providecommand{\BIBdecl}{\relax}
\BIBdecl

\bibitem{liu2020vision}
W.~Liu, L.~Xiong, X.~Xia, Y.~Lu, L.~Gao, and S.~Song, ``Vision-aided
  intelligent vehicle sideslip angle estimation based on a dynamic model,''
  \emph{IET Intelligent Transport Systems}, vol.~14, no.~10, pp. 1183--1189,
  2020.

\bibitem{liu2018intelligent}
W.~Liu, L.~Xiong, X.~Xia, and Z.~Yu, ``Intelligent vehicle sideslip angle
  estimation considering measurement signals delay,'' in \emph{2018 IEEE
  Intelligent Vehicles Symposium (IV)}.\hskip 1em plus 0.5em minus 0.4em\relax
  IEEE, 2018, pp. 1584--1589.

\bibitem{pal2021learning}
A.~Pal, Y.~Qiu, and H.~Christensen, ``Learning hierarchical relationships for
  object-goal navigation,'' in \emph{Conference on Robot Learning}.\hskip 1em
  plus 0.5em minus 0.4em\relax PMLR, 2021, pp. 517--528.

\bibitem{khalil2022licanet}
Y.~H. Khalil and H.~T. Mouftah, ``Licanet: Further enhancement of joint
  perception and motion prediction based on multi-modal fusion,'' \emph{IEEE
  Open Journal of Intelligent Transportation Systems}, vol.~3, pp. 222--235,
  2022.

\bibitem{kitajima2022nationwide}
S.~Kitajima, H.~Chouchane, J.~Antona-Makoshi, N.~Uchida, and J.~Tajima, ``A
  nationwide impact assessment of automated driving systems on traffic safety
  using multiagent traffic simulations,'' \emph{IEEE Open Journal of
  Intelligent Transportation Systems}, vol.~3, pp. 302--312, 2022.

\bibitem{papathanasopoulou2022data}
V.~Papathanasopoulou, I.~Spyropoulou, H.~Perakis, V.~Gikas, and
  E.~Andrikopoulou, ``A data-driven model for pedestrian behavior
  classification and trajectory prediction,'' \emph{IEEE Open Journal of
  Intelligent Transportation Systems}, vol.~3, pp. 328--339, 2022.

\bibitem{betz2022autonomous}
J.~Betz, H.~Zheng, A.~Liniger, U.~Rosolia, P.~Karle, M.~Behl, V.~Krovi, and
  R.~Mangharam, ``Autonomous vehicles on the edge: A survey on autonomous
  vehicle racing,'' \emph{IEEE Open Journal of Intelligent Transportation
  Systems}, vol.~3, pp. 458--488, 2022.

\bibitem{nayak2022uncertainty}
A.~Nayak, A.~Eskandarian, and Z.~Doerzaph, ``Uncertainty estimation of
  pedestrian future trajectory using bayesian approximation,'' \emph{IEEE Open
  Journal of Intelligent Transportation Systems}, vol.~3, pp. 617--630, 2022.

\bibitem{masmoudi2021reinforcement}
M.~Masmoudi, H.~Friji, H.~Ghazzai, and Y.~Massoud, ``A reinforcement learning
  framework for video frame-based autonomous car-following,'' \emph{IEEE Open
  Journal of Intelligent Transportation Systems}, vol.~2, pp. 111--127, 2021.

\bibitem{mukherjee2021predicting}
S.~Mukherjee, A.~M. Wallace, and S.~Wang, ``Predicting vehicle behavior using
  automotive radar and recurrent neural networks,'' \emph{IEEE Open Journal of
  Intelligent Transportation Systems}, vol.~2, pp. 254--268, 2021.

\bibitem{christensen2021autonomous}
H.~Christensen, D.~Paz, H.~Zhang, D.~Meyer, H.~Xiang, Y.~Han, Y.~Liu, A.~Liang,
  Z.~Zhong, and S.~Tang, ``Autonomous vehicles for micro-mobility,''
  \emph{Autonomous Intelligent Systems}, vol.~1, pp. 1--35, 2021.

\bibitem{de2021universally}
R.~De~Iaco, S.~L. Smith, and K.~Czarnecki, ``Universally safe swerve maneuvers
  for autonomous driving,'' \emph{IEEE Open Journal of Intelligent
  Transportation Systems}, vol.~2, pp. 482--494, 2021.

\bibitem{liu2022yolov5}
W.~Liu, K.~Quijano, and M.~M. Crawford, ``Yolov5-tassel: detecting tassels in
  rgb uav imagery with improved yolov5 based on transfer learning,'' \emph{IEEE
  Journal of Selected Topics in Applied Earth Observations and Remote Sensing},
  vol.~15, pp. 8085--8094, 2022.

\bibitem{han2021auto}
Y.~Han, Y.~Liu, D.~Paz, and H.~Christensen, ``Auto-calibration method using
  stop signs for urban autonomous driving applications,'' in \emph{2021 IEEE
  International Conference on Robotics and Automation (ICRA)}.\hskip 1em plus
  0.5em minus 0.4em\relax IEEE, 2021, pp. 13\,179--13\,185.

\bibitem{xu2021opencda}
R.~Xu, Y.~Guo, X.~Han, X.~Xia, H.~Xiang, and J.~Ma, ``Opencda: an open
  cooperative driving automation framework integrated with co-simulation,'' in
  \emph{2021 IEEE International Intelligent Transportation Systems Conference
  (ITSC)}.\hskip 1em plus 0.5em minus 0.4em\relax IEEE, 2021, pp. 1155--1162.

\bibitem{chen2023dynamic}
G.~Chen, X.~Zhao, Z.~Gao, and M.~Hua, ``Dynamic drifting control for general
  path tracking of autonomous vehicles,'' \emph{IEEE Transactions on
  Intelligent Vehicles}, 2023.

\bibitem{hua2019hierarchical}
M.~Hua, G.~Chen, B.~Zhang, and Y.~Huang, ``A hierarchical energy efficiency
  optimization control strategy for distributed drive electric vehicles,''
  \emph{Proceedings of the Institution of Mechanical Engineers, Part D: Journal
  of Automobile Engineering}, vol. 233, no.~3, pp. 605--621, 2019.

\bibitem{yu2022dair}
H.~Yu, Y.~Luo, M.~Shu, Y.~Huo, Z.~Yang, Y.~Shi, Z.~Guo, H.~Li, X.~Hu, J.~Yuan
  \emph{et~al.}, ``Dair-v2x: A large-scale dataset for vehicle-infrastructure
  cooperative 3d object detection,'' \emph{arXiv preprint arXiv:2204.05575},
  2022.

\bibitem{yuan2023generating}
Y.~Yuan, H.~Cheng, M.~Y. Yang, and M.~Sester, ``Generating evidential bev maps
  in continuous driving space,'' \emph{arXiv preprint arXiv:2302.02928}, 2023.

\bibitem{valiente2022robustness}
R.~Valiente, B.~Toghi, R.~Pedarsani, and Y.~P. Fallah, ``Robustness and
  adaptability of reinforcement learning-based cooperative autonomous driving
  in mixed-autonomy traffic,'' \emph{IEEE Open Journal of Intelligent
  Transportation Systems}, vol.~3, pp. 397--410, 2022.

\bibitem{ali2021optimizing}
Z.~Ali, W.~U. Khan, A.~Ihsan, O.~Waqar, G.~A.~S. Sidhu, and N.~Kumar,
  ``Optimizing resource allocation for 6g noma-enabled cooperative vehicular
  networks,'' \emph{IEEE Open Journal of Intelligent Transportation Systems},
  vol.~2, pp. 269--281, 2021.

\bibitem{isprs-archives-XLIII-B2-2021-255-2021}
Y.~Yuan and M.~Sester, ``Comap: A synthetic dataset for collective multi-agent
  perception of autonomous driving,'' \emph{The International Archives of the
  Photogrammetry, Remote Sensing and Spatial Information Sciences}, vol.
  XLIII-B2-2021, pp. 255--263, 2021.

\bibitem{xie2022safe}
H.~Xie, Y.~Wang, X.~Su, S.~Wang, and L.~Wang, ``Safe driving model based on v2v
  vehicle communication,'' \emph{IEEE Open Journal of Intelligent
  Transportation Systems}, vol.~3, pp. 449--457, 2022.

\bibitem{xu2022v2x}
R.~Xu, H.~Xiang, Z.~Tu, X.~Xia, M.-H. Yang, and J.~Ma, ``V2x-vit:
  Vehicle-to-everything cooperative perception with vision transformer,''
  \emph{arXiv preprint arXiv:2203.10638}, 2022.

\bibitem{hu2022where2comm}
Y.~Hu, S.~Fang, Z.~Lei, Y.~Zhong, and S.~Chen, ``Where2comm:
  Communication-efficient collaborative perception via spatial confidence
  maps,'' \emph{arXiv preprint arXiv:2209.12836}, 2022.

\bibitem{li2022v2x}
Y.~Li, D.~Ma, Z.~An, Z.~Wang, Y.~Zhong, S.~Chen, and C.~Feng, ``V2x-sim:
  Multi-agent collaborative perception dataset and benchmark for autonomous
  driving,'' \emph{IEEE Robotics and Automation Letters}, vol.~7, no.~4, pp.
  10\,914--10\,921, 2022.

\bibitem{yuan2022leveraging}
Y.~Yuan and M.~Sester, ``Leveraging dynamic objects for relative localization
  correction in a connected autonomous vehicle network,'' \emph{arXiv preprint
  arXiv:2205.09418}, 2022.

\bibitem{lei2022latency}
Z.~Lei, S.~Ren, Y.~Hu, W.~Zhang, and S.~Chen, ``Latency-aware collaborative
  perception,'' \emph{arXiv preprint arXiv:2207.08560}, 2022.

\bibitem{tan2021scenegen}
S.~Tan, K.~Wong, S.~Wang, S.~Manivasagam, M.~Ren, and R.~Urtasun, ``Scenegen:
  Learning to generate realistic traffic scenes,'' in \emph{Proceedings of the
  IEEE/CVF Conference on Computer Vision and Pattern Recognition}, 2021, pp.
  892--901.

\bibitem{suo2021trafficsim}
S.~Suo, S.~Regalado, S.~Casas, and R.~Urtasun, ``Trafficsim: Learning to
  simulate realistic multi-agent behaviors,'' in \emph{Proceedings of the
  IEEE/CVF Conference on Computer Vision and Pattern Recognition}, 2021, pp.
  10\,400--10\,409.

\bibitem{luo2020simulating}
Y.~Luo, P.~Cai, Y.~Lee, and D.~Hsu, ``Simulating autonomous driving in massive
  mixed urban traffic,'' \emph{arXiv preprint arXiv:2011.05767}, 2020.

\bibitem{abeysirigoonawardena2019generating}
Y.~Abeysirigoonawardena, F.~Shkurti, and G.~Dudek, ``Generating adversarial
  driving scenarios in high-fidelity simulators,'' in \emph{2019 International
  Conference on Robotics and Automation (ICRA)}.\hskip 1em plus 0.5em minus
  0.4em\relax IEEE, 2019, pp. 8271--8277.

\bibitem{rempe2022strive}
D.~Rempe, J.~Philion, L.~J. Guibas, S.~Fidler, and O.~Litany, ``Generating
  useful accident-prone driving scenarios via a learned traffic prior,'' in
  \emph{Conference on Computer Vision and Pattern Recognition (CVPR)}, 2022.

\bibitem{wang2021advsim}
J.~Wang, A.~Pun, J.~Tu, S.~Manivasagam, A.~Sadat, S.~Casas, M.~Ren, and
  R.~Urtasun, ``Advsim: Generating safety-critical scenarios for self-driving
  vehicles,'' in \emph{Proceedings of the IEEE/CVF Conference on Computer
  Vision and Pattern Recognition}, 2021, pp. 9909--9918.

\bibitem{liu2020who2com}
Y.-C. Liu, J.~Tian, C.-Y. Ma, N.~Glaser, C.-W. Kuo, and Z.~Kira, ``Who2com:
  Collaborative perception via learnable handshake communication,'' in
  \emph{2020 IEEE International Conference on Robotics and Automation
  (ICRA)}.\hskip 1em plus 0.5em minus 0.4em\relax IEEE, 2020, pp. 6876--6883.

\bibitem{dosovitskiy2017carla}
A.~Dosovitskiy, G.~Ros, F.~Codevilla, A.~Lopez, and V.~Koltun, ``Carla: An open
  urban driving simulator,'' in \emph{Conference on robot learning}.\hskip 1em
  plus 0.5em minus 0.4em\relax PMLR, 2017, pp. 1--16.

\bibitem{chen2019cooper}
Q.~Chen, S.~Tang, Q.~Yang, and S.~Fu, ``Cooper: Cooperative perception for
  connected autonomous vehicles based on 3d point clouds,'' in \emph{2019 IEEE
  39th International Conference on Distributed Computing Systems
  (ICDCS)}.\hskip 1em plus 0.5em minus 0.4em\relax OPTorganization, 2019, pp.
  514--524.

\bibitem{rawashdeh2018collaborative}
Z.~Y. Rawashdeh and Z.~Wang, ``Collaborative automated driving: A machine
  learning-based method to enhance the accuracy of shared information,'' in
  \emph{2018 21st International Conference on Intelligent Transportation
  Systems (ITSC)}.\hskip 1em plus 0.5em minus 0.4em\relax OPTorganization,
  2018, pp. 3961--3966.

\bibitem{rauch2012car2x}
A.~Rauch, F.~Klanner, R.~Rasshofer, and K.~Dietmayer, ``Car2x-based perception
  in a high-level fusion architecture for cooperative perception systems,'' in
  \emph{2012 IEEE Intelligent Vehicles Symposium}.\hskip 1em plus 0.5em minus
  0.4em\relax OPTorganization, 2012, pp. 270--275.

\bibitem{song2022vtc}
R.~Song, A.~Hegde, N.~Senel, A.~Knoll, and A.~Festag, ``Edge-aided sensor data
  sharing in vehicular communication networks,'' in \emph{2022 IEEE 95th
  Vehicular Technology Conference: (VTC2022-Spring)}, 2022, pp. 1--7.

\bibitem{wang2020v2vnet}
T.-H. Wang, S.~Manivasagam, M.~Liang, B.~Yang, W.~Zeng, and R.~Urtasun,
  ``V2vnet: Vehicle-to-vehicle communication for joint perception and
  prediction.''\hskip 1em plus 0.5em minus 0.4em\relax Springer, 2020, pp.
  605--621.

\bibitem{xu2022opv2v}
R.~Xu, H.~Xiang, X.~Xia, X.~Han, J.~Li, and J.~Ma, ``Opv2v: An open benchmark
  dataset and fusion pipeline for perception with vehicle-to-vehicle
  communication,'' in \emph{2022 International Conference on Robotics and
  Automation (ICRA)}.\hskip 1em plus 0.5em minus 0.4em\relax IEEE, 2022, pp.
  2583--2589.

\bibitem{yuan2022keypoints}
Y.~Yuan, H.~Cheng, and M.~Sester, ``Keypoints-based deep feature fusion for
  cooperative vehicle detection of autonomous driving,'' \emph{IEEE Robotics
  and Automation Letters}, 2022.

\bibitem{li2021learning}
Y.~Li, S.~Ren, P.~Wu, S.~Chen, C.~Feng, and W.~Zhang, ``Learning distilled
  collaboration graph for multi-agent perception,'' \emph{Advances in Neural
  Information Processing Systems}, vol.~34, 2021.

\bibitem{qiao2022adaptive}
D.~Qiao and F.~Zulkernine, ``Adaptive feature fusion for cooperative perception
  using lidar point clouds,'' \emph{arXiv preprint arXiv:2208.00116}, 2022.

\bibitem{xu2022cobevt}
R.~Xu, Z.~Tu, H.~Xiang, W.~Shao, B.~Zhou, and J.~Ma, ``Cobevt: Cooperative
  bird's eye view semantic segmentation with sparse transformers,'' \emph{arXiv
  preprint arXiv:2207.02202}, 2022.

\bibitem{cui2022coopernaut}
J.~Cui, H.~Qiu, D.~Chen, P.~Stone, and Y.~Zhu, ``Coopernaut: End-to-end driving
  with cooperative perception for networked vehicles,'' \emph{arXiv preprint
  arXiv:2205.02222}, 2022.

\bibitem{Su2022uncertainty}
S.~Su, Y.~Li, S.~He, S.~Han, C.~Feng, C.~Ding, and F.~Miao, ``Uncertainty
  quantification of collaborative detection for self-driving,'' 2023.

\bibitem{song2022federated}
R.~Song, L.~Zhou, V.~Lakshminarasimhan, A.~Festag, and A.~Knoll, ``Federated
  learning framework coping with hierarchical heterogeneity in cooperative
  its,'' \emph{arXiv preprint arXiv:2204.00215}, 2022.

\bibitem{song2022fedd3}
R.~Song, D.~Liu, D.~Z. Chen, A.~Festag, C.~Trinitis, M.~Schulz, and A.~Knoll,
  ``Federated learning via decentralized dataset distillation in
  resource-constrained edge environments,'' \emph{arXiv preprint
  arXiv:2208.11311}, 2022.

\bibitem{9234108}
G.~Chen, F.~Wang, W.~Li, L.~Hong, J.~Conradt, J.~Chen, Z.~Zhang, Y.~Lu, and
  A.~Knoll, ``Neuroiv: Neuromorphic vision meets intelligent vehicle towards
  safe driving with a new database and baseline evaluations,'' \emph{IEEE
  Transactions on Intelligent Transportation Systems}, vol.~23, no.~2, pp.
  1171--1183, 2022.

\bibitem{klischat2019generating}
M.~Klischat and M.~Althoff, ``Generating critical test scenarios for automated
  vehicles with evolutionary algorithms,'' in \emph{2019 IEEE Intelligent
  Vehicles Symposium (IV)}.\hskip 1em plus 0.5em minus 0.4em\relax IEEE, 2019,
  pp. 2352--2358.

\bibitem{norden2019efficient}
J.~Norden, M.~O'Kelly, and A.~Sinha, ``Efficient black-box assessment of
  autonomous vehicle safety,'' \emph{arXiv preprint arXiv:1912.03618}, 2019.

\bibitem{o2018scalable}
M.~O'Kelly, A.~Sinha, H.~Namkoong, R.~Tedrake, and J.~C. Duchi, ``Scalable
  end-to-end autonomous vehicle testing via rare-event simulation,''
  \emph{Advances in neural information processing systems}, vol.~31, 2018.

\bibitem{tu2020physically}
J.~Tu, M.~Ren, S.~Manivasagam, M.~Liang, B.~Yang, R.~Du, F.~Cheng, and
  R.~Urtasun, ``Physically realizable adversarial examples for lidar object
  detection,'' in \emph{Proceedings of the IEEE/CVF Conference on Computer
  Vision and Pattern Recognition}, 2020, pp. 13\,716--13\,725.

\bibitem{li2021fooling}
Y.~Li, C.~Wen, F.~Juefei-Xu, and C.~Feng, ``Fooling lidar perception via
  adversarial trajectory perturbation,'' in \emph{Proceedings of the IEEE/CVF
  International Conference on Computer Vision}, 2021, pp. 7898--7907.

\bibitem{cao2019adversarial}
Y.~Cao, C.~Xiao, B.~Cyr, Y.~Zhou, W.~Park, S.~Rampazzi, Q.~A. Chen, K.~Fu, and
  Z.~M. Mao, ``Adversarial sensor attack on lidar-based perception in
  autonomous driving,'' in \emph{Proceedings of the 2019 ACM SIGSAC conference
  on computer and communications security}, 2019, pp. 2267--2281.

\bibitem{tu2021adversarial}
J.~Tu, T.~Wang, J.~Wang, S.~Manivasagam, M.~Ren, and R.~Urtasun, ``Adversarial
  attacks on multi-agent communication,'' in \emph{Proceedings of the IEEE/CVF
  International Conference on Computer Vision}, 2021, pp. 7768--7777.

\bibitem{liu2020when2com}
Y.-C. Liu, J.~Tian, N.~Glaser, and Z.~Kira, ``When2com: Multi-agent perception
  via communication graph grouping,'' in \emph{Proceedings of the IEEE/CVF
  Conference on computer vision and pattern recognition}, 2020, pp. 4106--4115.

\bibitem{alzantot2019genattack}
M.~Alzantot, Y.~Sharma, S.~Chakraborty, H.~Zhang, C.-J. Hsieh, and M.~B.
  Srivastava, ``Genattack: Practical black-box attacks with gradient-free
  optimization,'' in \emph{Proceedings of the Genetic and Evolutionary
  Computation Conference}, 2019, pp. 1111--1119.

\bibitem{ru2019bayesopt}
B.~Ru, A.~Cobb, A.~Blaas, and Y.~Gal, ``Bayesopt adversarial attack,'' in
  \emph{International Conference on Learning Representations}, 2019.

\bibitem{lang2019pointpillars}
A.~H. Lang, S.~Vora, H.~Caesar, L.~Zhou, J.~Yang, and O.~Beijbom,
  ``Pointpillars: Fast encoders for object detection from point clouds,'' in
  \emph{Proceedings of the IEEE/CVF Conference on Computer Vision and Pattern
  Recognition}, 2019, pp. 12\,697--12\,705.

\bibitem{10045043}
R.~Xu, H.~Xiang, X.~Han, X.~Xia, Z.~Meng, C.-J. Chen, C.~Correa-Jullian, and
  J.~Ma, ``The opencda open-source ecosystem for cooperative driving automation
  research,'' \emph{IEEE Transactions on Intelligent Vehicles}, pp. 1--13,
  2023.

\end{thebibliography}
\end{document}